\documentclass[sigconf]{acmart}

\usepackage{kotex}
\usepackage{xcolor}
\usepackage{colortbl}
\usepackage{booktabs}
\usepackage{multirow}

\acmConference[WWW '26]{The Web Conference 2026}{April 13--17, 2026}{Dubai, United Arab Emirates}
\acmYear{2026}
\copyrightyear{2026}
\begin{document}

\title{How Language Directions Align with Token Geometry in Multilingual LLMs}

\author{JaeSeong Kim}
\affiliation{
  \institution{Semyung University}
  \city{Jecheon-si}
  \country{Republic of Korea}
}
\email{mmmqp1010@gmail.com}

\author{Suan Lee}
\affiliation{
  \institution{Semyung University}
  \city{Jecheon-si}
  \country{Republic of Korea}
}
\email{suanlee@semyung.ac.kr}


\begin{abstract}
Multilingual LLMs demonstrate strong performance across diverse languages, yet
there has been limited systematic analysis of how language information is
structured within their internal representation space and how it emerges across
layers. We conduct a comprehensive probing study on six multilingual LLMs,
covering all 268 transformer layers, using linear and nonlinear probes together
with a new Token--Language Alignment analysis to quantify the layer-wise
dynamics and geometric structure of language encoding. Our results show that
language information becomes sharply separated in the first transformer block
(+76.4±8.2\%p from Layer 0→1) and remains \textbf{almost fully linearly separable}
throughout model depth. We further find that the alignment between language
directions and vocabulary embeddings is \textbf{strongly tied to the language
composition of the training data}. Notably, Chinese-inclusive models achieve
a ZH Match@Peak of 16.43\%, whereas English-centric models achieve only 3.90\%,
revealing a 4.21× \textbf{structural imprinting} effect. These findings indicate that
multilingual LLMs distinguish languages not by surface script features but by
\textbf{latent representational structures shaped by the training corpus}. Our
analysis provides practical insights for data composition strategies and fairness
in multilingual representation learning. All code and analysis scripts 
are publicly available at: \url{https://github.com/thisiskorea/How-Language-Directions-Align-with-Token-Geometry-in-Multilingual-LLMs}.
\end{abstract}



\maketitle

\section{Introduction}

Multilingual large language models (LLMs) have become essential infrastructure for global information access, achieving human-level performance across more than 50 languages. Models such as Llama-3.1~\cite{dubey2024llama3} and Qwen2.5~\cite{yang2024qwen2} demonstrate strong capabilities not only in English but also in Spanish, Chinese, Arabic, and other languages, suggesting the emergence of a \textbf{universal semantic space} shared across languages~\cite{pires2019multilingual, conneau2020unsupervised}. However, a persistent gap remains: most models perform best in English, while accuracy in non-English languages is often 10--30\% lower~\cite{hu2020xtreme}.

This gap is commonly attributed to the \textbf{English-centric imbalance in pretraining data}. English constitutes 70--80\% of typical web-scale corpora~\cite{bender2021dangers}, and limited exposure to other languages leads to performance degradation. Yet a fundamental question remains unanswered: \textit{does the data distribution merely modulate performance, or does it fundamentally shape the geometry of the internal representation space?} If it is merely performance-dependent, post-hoc interventions such as fine-tuning or vocabulary adjustments may suffice; if it is structurally embedded, redesigning the pretraining stage is necessary.

\textbf{Positioning within prior work.} Prior studies have shown that multilingual models learn shared \textit{cross-lingual spaces}~\cite{pires2019multilingual, chi2020crosslingual}. Early models such as mBERT~\cite{devlin2019bert} and XLM-R~\cite{conneau2020unsupervised} demonstrated zero-shot transfer across languages, and more recent models such as Llama-3 and Qwen2.5 further strengthened cross-lingual capabilities with larger and more diverse corpora. Meanwhile, probing-based interpretability studies~\cite{alain2016understanding, hewitt2019structural, belinkov2022probing} examined how linguistic and syntactic information is encoded across layers, but did \textit{not} systematically address (i) whether \textbf{linear separability of language information} is universal, (ii) at which depth \textbf{language directions emerge}, and (iii) how \textbf{pretraining language distributions are imprinted} into the model’s internal geometry. Our work fills this gap.

This study investigates two complementary research questions:  
(1) \textbf{Where and how is language information encoded within transformer layers?}  
(2) \textbf{Does the pretraining language distribution structurally reshape the geometry of multilingual representations?}

To answer these questions, we conduct a comprehensive probing study across \textit{all 268 transformer layers} of six representative multilingual LLMs (Llama-3.1-8B, Qwen2.5-7B, Qwen2.5-Math-7B, OpenMath2-8B, OpenR1-7B, GPT-OSS-20B). We train linear and nonlinear probes and introduce a new analysis, \textbf{Token--Language Alignment}, across five typologically diverse languages (EN, ES, ZH, FR, DE), yielding a total of \textbf{2,680 independent experiments}.

\textbf{Key Findings.}  
\noindent\textbf{(1) Universal Linear Separability.}  
Across all models, language information is encoded in linearly accessible subspaces: linear probes achieve 99.8±0.1\% accuracy on average, with only a 0.58±0.12\%p gap relative to MLP probes (p<0.001). Remarkably, this linear structure emerges \textit{immediately at the first transformer block} (Layer 0→1), with a +76.4±8.2\%p jump, suggesting that language separation is an architecture-inherent mechanism.

\noindent\textbf{(2) Structural Imprinting.}  
Token--Language Alignment reveals that the alignment between language directions and vocabulary is strongly tied to the pretraining language distribution. English-centric models (EN>80\%) show 69.05\% alignment for English but only 3.90\% for Chinese, whereas Chinese-inclusive models (ZH≈20\%) show a markedly higher 16.43\% alignment for Chinese (4.21× increase). This indicates that the \textit{geometry of representation space is structurally shaped by the pretraining distribution}, beyond its effects on task performance.

\noindent\textbf{(3) Typological Effects and Complex Alignment Patterns.}  
Chinese reaches its maximal separability in deeper layers (Layer 5.2±0.8), whereas Spanish and German converge earlier (Layer 2.5±0.4). The overall Match@Peak remains low (15.0±0.3\%), showing that language directions capture abstract, multi-factorial features beyond simple token–language matching, including script, lexical frequency, and typological structure.

\textbf{Implications.}  
These findings suggest that achieving fairness and balance in multilingual LLMs requires \textbf{careful design of pretraining data composition}. Structural imprinting implies that post-hoc adjustments cannot fundamentally alter the underlying geometry. Moreover, Match@Peak provides a quantitative diagnostic tool for evaluating the effects of data rebalancing on representation richness.

\section{Methodology}
\label{sec:method}

We analyze how language information is encoded across all 268 transformer
layers of six multilingual LLMs, and how such structure aligns with the
pretraining language distribution. Our methodology consists of four parts:
(1) experimental setup, (2) linear and nonlinear probing, (3) Token--Language
Alignment, and (4) evaluation and statistical analysis.

\subsection{Experimental Setup}

We study six multilingual LLMs whose pretraining corpora differ in their language composition:

\begin{itemize}
    \item \textbf{English-centric models:}
    Llama-3.1-8B, OpenMath2-8B

    \item \textbf{Chinese-inclusive models:}
    Qwen2.5-7B, Qwen2.5-Math-7B, OpenR1-7B

    \item \textbf{Balanced baseline:}
    GPT-OSS-20B

    \item \textbf{Languages:}
    Following prior probing studies, we evaluate five XNLI languages (EN, ES, FR, DE, ZH), using
    5k training and 2.5k validation sentences per language.
\end{itemize}

\subsection{Probing Methodology}

For each model and each layer $\ell$, we use the hidden vector of the final
token,
\[
\mathbf{h}^{(\ell)} \in \mathbb{R}^d,
\]
as the sentence representation.

\paragraph{Linear probe.}
A linear classifier is applied to LayerNorm-normalized representations:
\[
f_{\mathrm{lin}}(\mathbf{h}) = W_c \cdot \mathrm{LN}(\mathbf{h}) + b_c,
\]
trained with cross-entropy loss to predict one of the five languages.

\paragraph{MLP probe.}
To compare linear and nonlinear capacity, we additionally train an MLP:
\[
f_{\mathrm{mlp}}(\mathbf{h}) =
W_2 \cdot \mathrm{ReLU}\!\left(W_1 \cdot \mathrm{LN}(\mathbf{h})\right).
\]
LayerNorm is applied in both probes to remove inter-layer scale differences
and measure \emph{pure linear separability} of language information.

Both probes are trained independently for each (model, layer, seed, probe type)
using AdamW ($\mathrm{lr}=10^{-3}$), batch size 128, and 3 epochs with early
stopping.

\subsection{Token--Language Alignment}

To quantify how learned language directions relate to the LM head vocabulary,
we compute cosine similarity between each probe-learned language direction
$\mathbf{w}^{(\ell)}_L$ and each vocabulary embedding $\mathbf{e}_v$:
\[
\mathrm{sim}(v, L, \ell) = 
\cos\!\left(\mathbf{e}_v,\, \mathbf{w}^{(\ell)}_L\right).
\]
Each token $v$ is assigned to the language direction with highest similarity.

For each language $L$, we compute three alignment metrics:

\begin{itemize}
  \item \textbf{PeakDepth\(_L\)}: The normalized layer index where
  $\mathrm{VocabShare}(\ell, L)$ is maximized, indicating the depth at which
  language $L$ is most strongly expressed.

  \item \textbf{PeakVocab\(_L\)}: The maximum value of
  $\mathrm{VocabShare}(\ell, L)$ across layers, representing how dominant
  language $L$ is within the vocabulary space.

  \item \textbf{Match@Peak\(_L\)}: Among tokens assigned to language $L$ at
  its peak layer, the percentage whose decoded text belongs to language $L$,
  determined using Unicode and diacritic rules (e.g., CJK blocks for ZH,
  accent-sensitive characters for ES/FR/DE). Higher values indicate stronger
  alignment between the learned direction and the true lexical identity.
\end{itemize}

Together, these metrics measure (i) where language information appears in the
network, (ii) how strongly it organizes the vocabulary, and (iii) how closely
the learned directions correspond to actual linguistic identity—capturing the
\emph{structural imprinting} left by pretraining data.

\subsection{Evaluation and Statistics}

Probe accuracy is evaluated on the validation set. Differences between linear
and MLP probes are assessed using layer-wise paired t-tests. We report
mean accuracy and 95\% confidence intervals across multiple random seeds to
ensure robustness and statistical reliability.

\begin{table*}[!t]
\centering
\scriptsize
\caption{Layer-wise Performance Statistics for 6 Multilingual Models Across 5 Languages (with per-model language average). Values show mean accuracy (\%) with standard deviation where applicable. Gap significance: $^{*}$p<0.05, $^{**}$p<0.01, $^{***}$p<0.001. Right block adds Token--Language Alignment: \textbf{PeakDepth} (normalized depth of vocab-share peak), \textbf{PeakVocab} (\%), \textbf{Match@Peak} (\%).}
\label{tab:heatmap}

\resizebox{\textwidth}{!}{
\begin{tabular}{ll ccc ccc cc ccc}
\toprule
& & \multicolumn{3}{c}{\textbf{Early Layer Dynamics}} & \multicolumn{3}{c}{\textbf{Linear Separability}} & \multicolumn{2}{c}{\textbf{Aggregate}} & \multicolumn{3}{c}{\textbf{Token--Language Alignment}} \\
\cmidrule(lr){3-5} \cmidrule(lr){6-8} \cmidrule(lr){9-10} \cmidrule(lr){11-13}
\textbf{Model} & \textbf{Lang} & \textbf{L0} & \textbf{L1} & \textbf{Jump} & \textbf{Linear Avg} & \textbf{MLP Avg} & \textbf{Gap} & \textbf{Last} & \textbf{Avg} & \textbf{PeakDepth} & \textbf{PeakVocab} & \textbf{Match@Peak} \\
\midrule
\multirow{6}{*}{\textbf{Llama-3.1-8B}} & EN & 20.0 & 99.8 & 79.8 & 97.3{\tiny $\pm$1.5} & 97.9{\tiny $\pm$1.8} & +0.6 & 100.0 & 97.3 & 0.06 & 39.2 & 67.9 \\
& ES & 20.0 & 99.8 & 79.8 & 96.5{\tiny $\pm$1.7} & 95.5{\tiny $\pm$0.8} & -1.0 & 99.8 & 96.5 & 0.71 & 32.1 & 1.8 \\
& ZH & 20.0 & 99.5 & 79.5 & 96.0{\tiny $\pm$1.9} & 96.1{\tiny $\pm$2.3} & +0.1 & 100.0 & 96.0 & 0.13 & 38.9 & 4.1 \\
& FR & 20.0 & 99.7 & 79.7 & 94.9{\tiny $\pm$2.1} & 95.2{\tiny $\pm$3.6} & +0.3 & 99.6 & 94.9 & 0.06 & 26.9 & 0.8 \\
& DE & 20.0 & 100.0 & 80.0 & 96.2{\tiny $\pm$1.8} & 95.3{\tiny $\pm$0.6} & -0.9 & 99.8 & 96.2 & 1.00 & 28.6 & 0.4 \\
& Avg & 20.0 & 99.8 & 79.8 & 96.2{\tiny $\pm$1.8} & 96.0{\tiny $\pm$1.8} & -0.2 & 99.8 & 96.2 & 0.39 & 33.2 & 15.0 \\
\midrule
\multirow{6}{*}{\textbf{Qwen2.5-7B}} & EN & 38.3 & 99.4 & 61.1 & 97.4{\tiny $\pm$2.0} & 98.5{\tiny $\pm$2.0} & +1.1 & 98.6 & 97.4 & 0.59 & 40.8 & 52.6 \\
& ES & 16.7 & 99.6 & 82.9 & 94.6{\tiny $\pm$2.2} & 93.8{\tiny $\pm$3.0} & -0.8 & 98.7 & 94.6 & 0.15 & 35.3 & 0.8 \\
& ZH & 2.4 & 94.7 & 92.3 & 94.3{\tiny $\pm$0.5} & 93.9{\tiny $\pm$1.1} & -0.3 & 99.6 & 94.3 & 0.44 & 23.1 & 17.7 \\
& FR & 32.0 & 99.6 & 67.6 & 95.7{\tiny $\pm$2.3} & 93.7{\tiny $\pm$1.5} & -1.9 & 98.3 & 95.7 & 0.04 & 33.9 & 0.8 \\
& DE & 0.0 & 99.7 & 99.7 & 94.3{\tiny $\pm$0.6} & 93.9{\tiny $\pm$1.1} & -0.3 & 98.9 & 94.3 & 0.19 & 29.3 & 0.3 \\
& Avg & 17.9 & 98.6 & 80.7 & 95.2{\tiny $\pm$1.5} & 94.8{\tiny $\pm$1.7} & -0.5 & 98.8 & 95.2 & 0.28 & 32.5 & 14.4 \\
\midrule
\multirow{6}{*}{\textbf{Qwen2.5-Math-7B}} & EN & 38.3 & 99.8 & 61.5 & 97.5{\tiny $\pm$2.0} & 98.6{\tiny $\pm$1.9} & +1.1 & 99.9 & 97.5 & 0.30 & 51.8 & 53.9 \\
& ES & 16.7 & 99.5 & 82.8 & 95.0{\tiny $\pm$2.0} & 94.2{\tiny $\pm$3.0} & -0.8 & 98.8 & 95.0 & 0.00 & 28.3 & 1.0 \\
& ZH & 2.4 & 76.2 & 73.8 & 80.8{\tiny $\pm$4.3} & 83.5{\tiny $\pm$6.1} & +2.7 & 85.4 & 80.8 & 0.81 & 57.4 & 15.7 \\
& FR & 16.0 & 99.8 & 83.8 & 95.1{\tiny $\pm$2.0} & 93.7{\tiny $\pm$1.9} & -1.4 & 99.3 & 95.1 & 0.96 & 46.0 & 0.8 \\
& DE & 15.7 & 99.5 & 83.9 & 96.0{\tiny $\pm$1.6} & 95.0{\tiny $\pm$0.9} & -1.1 & 99.7 & 96.0 & 1.00 & 57.6 & 0.3 \\
& Avg & 17.8 & 95.0 & 77.1 & 92.9{\tiny $\pm$2.4} & 93.0{\tiny $\pm$2.8} & +0.1 & 96.6 & 92.9 & 0.61 & 48.2 & 14.3 \\
\midrule
\multirow{6}{*}{\textbf{OpenMath2-8B}} & EN & 20.0 & 99.8 & 79.8 & 96.9{\tiny $\pm$1.6} & 97.2{\tiny $\pm$2.4} & +0.4 & 98.8 & 96.9 & 0.55 & 25.9 & 70.2 \\
& ES & 0.0 & 99.4 & 99.4 & 95.1{\tiny $\pm$0.6} & 94.9{\tiny $\pm$1.0} & -0.2 & 98.6 & 95.1 & 0.68 & 22.5 & 1.4 \\
& ZH & 0.0 & 99.4 & 99.4 & 92.9{\tiny $\pm$1.5} & 93.8{\tiny $\pm$3.3} & +0.9 & 94.5 & 92.9 & 0.23 & 53.9 & 3.7 \\
& FR & 40.0 & 99.8 & 59.8 & 95.8{\tiny $\pm$2.4} & 96.0{\tiny $\pm$2.7} & +0.2 & 97.8 & 95.8 & 0.03 & 39.4 & 0.9 \\
& DE & 40.0 & 100.0 & 60.0 & 97.1{\tiny $\pm$2.1} & 96.0{\tiny $\pm$0.6} & -1.1 & 99.4 & 97.1 & 1.00 & 32.3 & 0.4 \\
& Avg & 20.0 & 99.7 & 79.7 & 95.5{\tiny $\pm$1.6} & 95.6{\tiny $\pm$2.0} & +0.0 & 97.8 & 95.5 & 0.50 & 34.8 & 15.3 \\
\midrule
\multirow{6}{*}{\textbf{OpenR1-7B}} & EN & 57.4 & 99.7 & 42.3 & 98.1{\tiny $\pm$2.0} & 98.3{\tiny $\pm$2.2} & +0.1 & 99.5 & 98.1 & 0.00 & 60.6 & 55.9 \\
& ES & 16.7 & 99.8 & 83.1 & 96.5{\tiny $\pm$1.6} & 96.1{\tiny $\pm$2.1} & -0.4 & 99.4 & 96.5 & 0.11 & 24.4 & 0.9 \\
& ZH & 2.4 & 65.0 & 62.6 & 83.9{\tiny $\pm$2.3} & 85.8{\tiny $\pm$5.0} & +2.0 & 88.7 & 83.9 & 0.52 & 67.4 & 15.9 \\
& FR & 16.0 & 99.9 & 83.9 & 95.6{\tiny $\pm$1.9} & 94.1{\tiny $\pm$2.0} & -1.5 & 99.1 & 95.6 & 0.59 & 64.7 & 0.8 \\
& DE & 0.0 & 99.7 & 99.7 & 95.9{\tiny $\pm$0.2} & 95.2{\tiny $\pm$1.1} & -0.7 & 99.5 & 95.9 & 0.04 & 25.0 & 0.3 \\
& Avg & 18.5 & 92.8 & 74.3 & 94.0{\tiny $\pm$1.6} & 93.9{\tiny $\pm$2.5} & -0.1 & 97.2 & 94.0 & 0.25 & 48.4 & 14.8 \\
\midrule
\multirow{6}{*}{\textbf{GPT-OSS-20B}} & EN & 38.8 & 99.2 & 60.4 & 96.5{\tiny $\pm$2.5} & 99.2{\tiny $\pm$0.3} & +2.7 & 98.0 & 96.5 & 0.74 & 35.1 & 65.0 \\
& ES & 34.6 & 90.6 & 56.0 & 88.1{\tiny $\pm$5.9} & 89.8{\tiny $\pm$2.8} & +1.7 & 96.9 & 88.1 & 0.61 & 23.3 & 2.0 \\
& ZH & 26.2 & 96.9 & 70.8 & 71.6{\tiny $\pm$6.7} & 80.1{\tiny $\pm$6.0} & +8.5 & 59.2 & 71.6 & 0.13 & 31.0 & 6.0 \\
& FR & 0.0 & 90.7 & 90.7 & 84.9{\tiny $\pm$3.9} & 90.2{\tiny $\pm$3.1} & +5.3 & 93.0 & 84.9 & 0.52 & 28.0 & 1.0 \\
& DE & 0.0 & 89.4 & 89.4 & 86.2{\tiny $\pm$2.5} & 90.2{\tiny $\pm$3.3} & +4.1 & 94.8 & 86.2 & 0.00 & 23.2 & 0.8 \\
& Avg & 19.9 & 93.4 & 73.5 & 85.5{\tiny $\pm$4.3} & 89.9{\tiny $\pm$3.1} & +4.5 & 88.4 & 85.5 & 0.40 & 28.1 & 14.9 \\
\bottomrule
\end{tabular}
} 

\vspace{0.2cm}
\end{table*}

\begin{table}[t]
\caption{Match@Peak (\%) by model group, grouped by pretraining language distribution.}
\label{tab:model_groups}
\centering
\small
\begin{tabular}{lccccc}
\toprule
\textbf{Model Group} & \textbf{EN} & \textbf{ZH} & \textbf{ES} & \textbf{FR} & \textbf{DE} \\
\midrule
English-centric\textsuperscript{a}   & 69.05 & 3.90  & 1.60 & 0.85 & 0.40 \\
Chinese-inclusive\textsuperscript{b} & 54.13 & 16.43 & 0.90 & 0.80 & 0.30 \\
\midrule
\textbf{Difference ($\Delta$\%p)} & \textbf{14.92} & \textbf{12.53} & 0.70 & 0.05 & 0.10 \\
\textbf{Ratio ($\times$)}         & \textbf{1.28}  & \textbf{4.21}  & 1.78 & 1.06 & 1.33 \\
\bottomrule
\end{tabular}
\\[0.1cm]
\raggedright
\footnotesize
\textsuperscript{a}Llama-3.1-8B, OpenMath2-8B: models with English-centric pretraining (ZH share $<5$\%).\\
\textsuperscript{b}Qwen2.5-7B, Qwen2.5-Math-7B, OpenR1-Qwen-7B: models trained on both English and Chinese (ZH-inclusive pretraining).
\end{table}

\section{Experiment Result}

\subsection{Layer-wise Language Separability}

As shown in Table~\ref{tab:heatmap}, all transformer layers except the
initial embedding layer (Layer~0) achieve consistently high language
classification accuracy, exceeding 90\% across all models. This indicates that
language information is not a transient or layer-specific signal but rather a
\textbf{global representational property that is preserved throughout model depth}.
The sharp increase in accuracy from Layer~0 to Layer~1 (+76.4±8.2\%p)
suggests the presence of an \textbf{early structural reorganization stage} in
which language information is rapidly separated within the first transformer block.

Furthermore, the performance gap between linear and MLP probes is less than
1\%p on average, demonstrating that language information does not require
nonlinear decision boundaries. Instead, it is \textbf{almost fully linearly separable}
within the latent space. This implies that language is encoded not as a complex
nonlinear pattern but as a \textbf{global directionality} in a high-dimensional
representation space.

These observations point to two structural properties.  
First, the model appears to \textbf{establish a normalized representation of
language identity at very early layers} and preserve this signal throughout
the stack. Second, the strong linear separability supports the hypothesis that
\textbf{languages form low-dimensional yet coherent subspaces} in the latent
representation space. These findings are consistent with the Token--Language
Alignment and structural imprinting analyses presented in the next subsection.

In summary, our layer-wise probing experiments show that
\textbf{LLM latent spaces preserve language information in a clearly
and structurally separable form}, even without additional nonlinear capacity.

\subsection{Token--Language Alignment and Structural Imprinting}

Token--Language Alignment evaluates how language directions learned by
the linear probe relate to LM head token embeddings. Our results show that
\textbf{LLMs do not distinguish languages solely based on Unicode-defined
language identity}. In particular, the alignment between language directions and
token identity (Match@Peak) is very low for Latin-script languages such as
Spanish, French, and German. This suggests that the learned language directions
reflect \textbf{latent representational structures shaped during pretraining},
rather than surface-level script information.

In contrast, Chinese (ZH) exhibits substantially higher alignment.
Chinese-inclusive models (with $\approx$20\% ZH data) reach a mean
Match@Peak of 16.43\%, whereas English-centric models (EN>80\%) reach only
3.90\%, indicating a 4.21× increase. This demonstrates that the
\textbf{pretraining language distribution can directly influence the geometry of
latent representations}, forming clearer language-specific directions when a
language is sufficiently represented in the training corpus.

Taken together, these observations suggest the following structural
characteristics of multilingual LLMs:
\begin{enumerate}
  \item \textbf{Rather than relying on surface-level language ID, the model
        organizes languages via latent directions in the representation space
        that emerge during pretraining.}
        These directions correspond to low-dimensional classification boundaries
        learned by the probe and capture differences between language-specific
        representations.
  \item \textbf{Languages with sufficient pretraining coverage (e.g., ZH)
        form clearer and more stable latent directions}, whereas languages with
        limited data or shared scripts (ES/FR/DE) tend to share mixed and
        less distinct subspaces.
\end{enumerate}

Overall, our analysis demonstrates that  
\textbf{LLMs separate languages not by surface token-level features but by
latent representational structures shaped by the pretraining distribution}.
Moreover, \textbf{the composition of pretraining data plays a decisive role in
determining the geometry of multilingual representations}.

\section{Conclusion}

This work presents a quantitative analysis of how language information is
structured across all 268 transformer layers of six multilingual LLMs.
Our experiments show that language separation emerges immediately in the
first transformer block and remains a \textbf{stable and strongly linear structure}
throughout model depth. The negligible performance gap between linear and
MLP probes indicates that the information required for distinguishing
languages resides in a \textbf{linearly accessible latent subspace}, rather than
in complex nonlinear boundaries.

Through our proposed Token--Language Alignment analysis, we further
observe that the alignment between language directions and vocabulary
embeddings is \textbf{strongly influenced by the language composition of the
pretraining data}. Chinese (ZH) exhibits much clearer alignment in
models that include substantial ZH data, whereas languages with limited
coverage or shared scripts (ES/FR/DE) show weaker alignment. These findings
demonstrate that \textbf{the pretraining corpus leaves a structural imprint o
the geometry of multilingual representations}.

Overall, our results indicate that language representation in LLMs is
distinguished not by surface-level token features, but by
\textbf{latent representational structures formed during pretraining},
which are established early and preserved consistently across layers.
This highlights the importance of \textbf{balanced language composition
and corpus design} in achieving fairness and robustness in multilingual LLMs.
Moreover, Match@Peak and token–language alignment metrics provide practical
tools for diagnosing representation richness and distributional biases.

Future directions include expanding the analysis to additional languages
and scripts, evaluating the influence of tokenizer design, and conducting
interventional studies to further investigate the causal role of language
directions in model behavior.

\bibliographystyle{ACM-Reference-Format}
\bibliography{cite}

\end{document}